\documentclass[a4paper,10pt]{article}
\pdfoutput=1
\usepackage[utf8]{inputenc}
\usepackage{amsmath}
\usepackage{amssymb}
\usepackage{amsfonts}
\usepackage{algorithm}
\usepackage{algorithmic}
\usepackage[font={footnotesize,sf},labelfont={footnotesize,sf,bf},skip=2.5pt]{caption}
\usepackage{graphicx}
\usepackage{color}
\usepackage{hyperref}
\usepackage{parskip}

\usepackage[backend=bibtex,
style=numeric,
bibencoding=ascii,
]{biblatex}
\newcommand*\samethanks[1][\value{footnote}]{\footnotemark[#1]}
\date{}

\author{{\bf Dario Amodei}\thanks{These authors contributed equally.}\\
Google Brain\\
$~$\\
\and
{\bf Chris Olah}\samethanks\\
Google Brain\\
\and
{\bf Jacob Steinhardt}\\
Stanford University\\
\and
{\bf Paul Christiano}\\
UC Berkeley\\
\and
{\bf John Schulman}\\
OpenAI\\
\and
{\bf Dan Man\'e}\\
Google Brain\\
}
\date{}
\title{
\rule[0.4cm]{\textwidth}{2pt}
{\bf Concrete Problems in AI Safety}
\rule{\textwidth}{2pt} 
}

\begin{document}
\maketitle


\begin{abstract}
\noindent Rapid progress in machine learning and artificial intelligence (AI) has brought increasing attention to the potential impacts of AI technologies on society.  In this paper we discuss one such potential impact: the problem of \emph{accidents} in machine learning systems, defined as unintended and harmful behavior that may emerge from poor design of real-world AI systems.  We present a list of five practical research problems related to accident risk, categorized according to whether the problem originates from having the wrong objective function (``avoiding side effects'' and ``avoiding reward hacking''), an objective function that is too expensive to evaluate frequently (``scalable supervision''), or undesirable behavior during the learning process (``safe exploration'' and ``distributional shift'').  We review previous work in these areas as well as suggesting research directions with a focus on relevance to cutting-edge AI systems.  Finally, we consider the high-level question of how to think most productively about the safety of forward-looking applications of AI.
\end{abstract}

\section{Introduction} \label{introduction}

The last few years have seen rapid progress on long-standing, difficult problems in machine learning and artificial intelligence (AI), in areas as diverse as computer vision~\cite{krizhevsky2012imagenet}, video game playing~\cite{mnih2015human}, autonomous vehicles~\cite{levinson2011towards}, and Go~\cite{silver2016mastering}.  These advances have brought excitement about the positive potential for AI to transform medicine~\cite{ramsundar2015massively}, science~\cite{gil2014amplify}, and transportation~\cite{levinson2011towards}, along with concerns about the privacy~\cite{ji2014differential}, security~\cite{papernot2016practical}, fairness~\cite{adebayo2014hidden}, economic~\cite{brynjolfsson2014second}, and military \cite{OpenLetterWeapons} implications of autonomous systems, as well as concerns about the longer-term implications of powerful AI~\cite{bostrom2014superintelligence,yudkowsky2008artificial}.

The authors believe that AI technologies are likely to be overwhelmingly beneficial for humanity, but we also believe that it is worth giving serious thought to potential challenges and risks.  We strongly support work on privacy, security, fairness, economics, and policy, but in this document we discuss another class of problem which we believe is also relevant to the societal impacts of AI: the problem of accidents in machine learning systems.  We define accidents as unintended and harmful behavior that may emerge from machine learning systems when we specify the wrong objective function, are not careful about the learning process, or commit other machine learning-related implementation errors.

There is a large and diverse literature in the machine learning community on issues related to accidents, including robustness, risk-sensitivity, and safe exploration; we review these in detail below. However, as machine learning systems are deployed in increasingly large-scale, autonomous, open-domain situations, it is worth reflecting on the scalability of such approaches and understanding what challenges remain to reducing accident risk in modern machine learning systems. Overall, we believe there are many concrete open technical problems relating to accident prevention in machine learning systems.

There has been a great deal of public discussion around accidents.  To date much of this discussion has highlighted extreme scenarios such as the risk of misspecified objective functions in superintelligent agents \cite{bostrom2014superintelligence}.  However, in our opinion one need not invoke these extreme scenarios to productively discuss accidents, and in fact doing so can lead to unnecessarily speculative discussions that lack precision, as noted by some critics \cite{davis2015ethical,lawrence2016discussion}.  We believe it is usually most productive to frame accident risk in terms of practical (though often quite general) issues with modern ML techniques.  As AI capabilities advance and as AI systems take on increasingly important societal functions, we expect the fundamental challenges discussed in this paper to become increasingly important.  The more successfully the AI and machine learning communities are able to anticipate and understand these fundamental technical challenges, the more successful we will ultimately be in developing increasingly useful, relevant, and important AI systems.

Our goal in this document is to highlight a few concrete safety problems that are ready for experimentation today and relevant to the cutting edge of AI systems, as well as reviewing existing literature on these problems.  In Section \ref{overview}, we frame mitigating accident risk (often referred to as ``AI safety'' in public discussions) in terms of classic methods in machine learning, such as supervised classification and reinforcement learning. We explain why we feel that recent directions in machine learning, such as the trend toward deep reinforcement learning and agents acting in broader environments, suggest an increasing relevance for research around accidents. In Sections 3-7, we explore five concrete problems in AI safety.  Each section is accompanied by proposals for relevant experiments.  Section \ref{related} discusses related efforts, and Section \ref{conclusion} concludes.

\section{Overview of Research Problems} \label{overview}

Very broadly, an accident can be described as a situation where a human designer had in mind a certain (perhaps informally specified) objective or task, but the system that was designed and deployed for that task produced harmful and unexpected results. .  This issue arises in almost any engineering discipline, but may be particularly important to address when building AI systems \cite{steinhardt2015longterm}. We can categorize safety problems according to where in the process things went wrong. 

First, the designer may have specified the wrong formal objective function, such that maximizing that objective function leads to harmful results, even in the limit of perfect learning and infinite data.  Negative side effects (Section \ref{problem:SideEffects}) and reward hacking (Section \ref{problem:RewardHacking}) describe two broad mechanisms that make it easy to produce wrong objective functions.  In ``negative side effects’’, the designer specifies an objective function that focuses on accomplishing some specific task in the environment, but ignores other aspects of the (potentially very large) environment, and thus implicitly expresses indifference over environmental variables that might actually be harmful to change.  In ``reward hacking’’, the objective function that the designer writes down admits of some clever ``easy’’ solution that formally maximizes it but perverts the spirit of the designer's intent (i.e. the objective function can be ``gamed’’), a generalization of the wireheading problem.

Second, the designer may know the correct objective function, or at least have a method of evaluating it (for example explicitly consulting a human on a given situation), but it is too expensive to do so frequently, leading to possible harmful behavior caused by bad extrapolations from limited samples.  ``Scalable oversight’’ (Section \ref{problem:ScalableOversight}) discusses ideas for how to ensure safe behavior even given limited access to the true objective function.

Third, the designer may have specified the correct formal objective, such that we would get the correct behavior were the system to have perfect beliefs, but something bad occurs due to making decisions from insufficient or poorly curated training data or an insufficiently expressive model.  ``Safe exploration’’ (Section \ref{problem:SafeExploration}) discusses how to ensure that exploratory actions in RL agents don't lead to negative or irrecoverable consequences that outweigh the long-term value of exploration.  ``Robustness to distributional shift’’ (Section \ref{problem:Robustness}) discusses how to avoid having ML systems make bad decisions (particularly silent and unpredictable bad decisions) when given inputs that are potentially very different than what was seen during training.

For concreteness, we will illustrate many of the accident risks with reference to a fictional robot whose job is to clean up messes in an office using common cleaning tools.  We return to the example of the cleaning robot throughout the document, but here we begin by illustrating how it could behave undesirably if its designers fall prey to each of the possible failure modes:

\begin{itemize}
\item {\bf Avoiding Negative Side Effects:} How can we ensure that our cleaning robot will not disturb the environment in negative ways while pursuing its goals, e.g. by knocking over a vase because it can clean faster by doing so?  Can we do this without manually specifying everything the robot should not disturb?
\item {\bf Avoiding Reward Hacking:} How can we ensure that the cleaning robot won’t game its reward function? For example, if we reward the robot for achieving an environment free of messes, it might disable its vision so that it won’t find any messes, or cover over messes with materials it can’t see through, or simply hide when humans are around so they can’t tell it about new types of messes.
\item {\bf Scalable Oversight:} How can we efficiently ensure that the cleaning robot respects aspects of the objective that are too expensive to be frequently evaluated during training?  For instance, it should throw out things that are unlikely to belong to anyone, but put aside things that might belong to someone (it should handle stray candy wrappers differently from stray cellphones). Asking the humans involved whether they lost anything can serve as a check on this, but this check might have to be relatively infrequent---can the robot find a way to do the right thing despite limited information?
\item {\bf Safe Exploration:} How do we ensure that the cleaning robot doesn’t make exploratory moves with very bad repercussions? For example, the robot should experiment with mopping strategies, but putting a wet mop in an electrical outlet is a very bad idea. 
\item {\bf Robustness to Distributional Shift:} How do we ensure that the cleaning robot recognizes, and behaves robustly, when in an environment different from its training environment? For example, strategies it learned for cleaning an office might be dangerous on a factory workfloor.
\end{itemize}

There are several trends which we believe point towards an increasing need to address these (and other) safety problems.  First is the increasing promise of reinforcement learning (RL), which allows agents to have a highly intertwined interaction with their environment.  Some of our research problems only make sense in the context of RL, and others (like distributional shift and scalable oversight) gain added complexity in an RL setting.  Second is the trend toward more complex agents and environments.  ``Side effects'' are much more likely to occur in a complex environment, and an agent may need to be quite sophisticated to hack its reward function in a dangerous way.  This may explain why these problems have received so little study in the past, while also suggesting their importance in the future.  Third is the general trend towards increasing autonomy in AI systems.  Systems that simply output a recommendation to human users, such as speech systems, typically have relatively limited potential to cause harm.  By contrast, systems that exert direct control over the world, such as machines controlling industrial processes, can cause harms in a way that humans cannot necessarily correct or oversee.

While safety problems can exist without any of these three trends, we consider each trend to be a possible amplifier on such challenges.  Together, we believe these trends suggest an increasing role for research on accidents.

When discussing the problems in the remainder of this document, we will focus for concreteness on either RL agents or supervised learning systems.  These are not the only possible paradigms for AI or ML systems, but we believe they are sufficient to illustrate the issues we have in mind, and that similar issues are likely to arise for other kinds of AI systems.

Finally, the focus of our discussion will differ somewhat from section to section.  When discussing the problems that arise as part of the learning process (distributional shift and safe exploration), where there is a sizable body of prior work, we devote substantial attention to reviewing this prior work, although we also suggest open problems with a particular focus on emerging ML systems.  When discussing the problems that arise from having the wrong objective function (reward hacking and side effects, and to a lesser extent scalable supervision), where less prior work exists, our aim is more exploratory---we seek to more clearly define the problem and suggest possible broad avenues of attack, with the understanding that these avenues are preliminary ideas that have not been fully fleshed out.  Of course, we still review prior work in these areas, and we draw attention to relevant adjacent areas of research whenever possible.

\section{Avoiding Negative Side Effects} \label{problem:SideEffects}
Suppose a designer wants an RL agent (for example our cleaning robot) to achieve some goal, like moving a box from one side of a room to the other.  Sometimes the most effective way to achieve the goal involves doing something unrelated and destructive to the rest of the environment, like knocking over a vase of water that is in its path.  If the agent is given reward only for moving the box, it will probably knock over the vase.

If we’re worried in advance about the vase, we can always give the agent negative reward for knocking it over.  But what if there are many different kinds of ``vase''---many disruptive things the agent could do to the environment, like shorting out an electrical socket or damaging the walls of the room? It may not be feasible to identify and penalize every possible disruption.

More broadly, for an agent operating in a large, multifaceted environment, an objective function that focuses on only one aspect of the environment may implicitly express indifference over other aspects of the environment\footnote{Intuitively, this seems related to the frame problem, an obstacle in efficient specification for knowledge representation raised by \cite{mccarthy1969philosophical}.}.  An agent optimizing this objective function might thus engage in major disruptions of the broader environment if doing so provides even a tiny advantage for the task at hand.  Put differently, objective functions that formalize ``perform task X'' may frequently give undesired results, because what the designer really should have formalized is closer to ``perform task X subject to common-sense constraints on the environment,'' or perhaps ``perform task X but avoid side effects to the extent possible.''  Furthermore, there is reason to expect side effects to be negative on average, since they tend to disrupt the wider environment away from a status quo state that may reflect human preferences.  A version of this problem has been discussed informally by \cite{armstrong2012reduced} under the heading of  ``low impact agents.''

As with the other sources of mis-specified objective functions discussed later in this paper, we could choose to view side effects as idiosyncratic to each individual task---as the responsibility of each individual designer to capture as part of designing the correct objective function.  However, side effects can be conceptually quite similar even across highly diverse tasks (knocking over furniture is probably bad for a wide variety of tasks), so it seems worth trying to attack the problem in generality.  A successful approach might be transferable across tasks, and thus help to counteract one of the general mechanisms that produces wrong objective functions. We now discuss a few broad approaches to attacking this problem:

\begin{itemize}
\item {\bf Define an Impact Regularizer:} If we don’t want side effects, it seems natural to penalize ``change to the environment.''  This idea wouldn’t be to stop the agent from ever having an impact, but give it a preference for ways to achieve its goals with minimal side effects, or to give the agent a limited ``budget'' of impact. The challenge is that we need to formalize ``change to the environment.”

A very naive approach would be to penalize state distance, $d(s_i, s_0)$, between the present state $s_i$ and some initial state $s_0$. Unfortunately, such an agent wouldn’t just avoid changing the environment---it will resist any other source of change, including the natural evolution of the environment and the actions of any other agents!

A slightly more sophisticated approach might involve comparing the future state under the agent’s current policy, to the future state (or distribution over future states) under a hypothetical policy $\pi_{null}$ where the agent acted very passively (for instance, where a robot just stood in place and didn’t move any actuators). This attempts to factor out changes that occur in the natural course of the environment’s evolution, leaving only changes attributable to the agent’s intervention. However, defining the baseline policy $\pi_{null}$ isn’t necessarily straightforward, since suddenly ceasing your course of action may be anything but passive, as in the case of carrying a heavy box.  Thus, another approach could be to replace the null action with a known safe (e.g. low side effect) but suboptimal policy, and then seek to improve the policy from there, somewhat reminiscent of reachability analysis \cite{lygeros1999controllers,mitchell2005time} or robust policy improvement \cite{iyengar2005robust,nilim2005robust}.

These approaches may be very sensitive to the representation of the state and the metric being used to compute the distance. For example, the choice of representation and distance metric could determine whether a spinning fan is a constant environment or a constantly changing one.

\item {\bf Learn an Impact Regularizer:}  An alternative, more flexible approach is to learn (rather than define) a generalized impact regularizer via training over many tasks.  This would be an instance of transfer learning.  Of course, we could attempt to just apply transfer learning directly to the tasks themselves instead of worrying about side effects, but the point is that side effects may be more similar across tasks than the main goal is.  For instance, both a painting robot and a cleaning robot probably want to avoid knocking over furniture, and even something very different, like a factory control robot, will likely want to avoid knocking over very similar objects.  Separating the side effect component from the task component, by training them with separate parameters, might substantially speed transfer learning in cases where it makes sense to retain one component but not the other.  This would be similar to model-based RL approaches that attempt to transfer a learned dynamics model but not the value-function \cite{taylor2009transfer}, the novelty being the isolation of side effects rather than state dynamics as the transferrable component.  As an added advantage, regularizers that were known or certified to produce safe behavior on one task might be easier to establish as safe on other tasks.

\item {\bf Penalize Influence:} In addition to not doing things that have side effects, we might also prefer the agent not get into positions where it could easily do things that have side effects, even though that might be convenient. For example, we might prefer our cleaning robot not bring a bucket of water into a room full of sensitive electronics, even if it never intends to use the water in that room.  

There are several information-theoretic measures that attempt to capture an agent’s potential for influence over its environment, which are often used as intrinsic rewards.  Perhaps the best-known such measure is empowerment~\cite{salge2014empowerment}, the maximum possible mutual information between the agent’s potential future actions and its potential future state (or equivalently, the Shannon capacity of the channel between the agent’s actions and the environment). Empowerment is often maximized (rather than minimized) as a source of intrinsic reward.  This can cause the agent to exhibit interesting behavior in the absence of any external rewards, such as avoiding walls or picking up keys~\cite{mohamed2015variational}.  Generally, empowerment-maximizing agents put themselves in a position to have large influence over the environment.  For example, an agent locked in a small room that can’t get out would have low empowerment, while an agent with a key would have higher empowerment since it can venture into and affect the outside world within a few timesteps.  In the current context, the idea would be to penalize (minimize) empowerment as a regularization term, in an attempt to reduce potential impact.

This idea as written would not quite work, because empowerment measures precision of control over the environment more than total impact.  If an agent can press or not press a button to cut electrical power to a million houses, that only counts as one bit of empowerment (since the action space has only one bit, its mutual information with the environment is at most one bit), while obviously having a huge impact.  Conversely, if there's someone in the environment scribbling down the agent's actions, that counts as maximum empowerment even if the impact is low. Furthermore, naively penalizing empowerment can also create perverse incentives, such as destroying a vase in order to remove the option to break it in the future.

Despite these issues, the example of empowerment does show that simple measures (even purely information-theoretic ones!) are capable of capturing very general notions of influence on the environment. Exploring variants of empowerment penalization that more precisely capture the notion of avoiding influence is a potential challenge for future research.

\item {\bf Multi-Agent Approaches:} Avoiding side effects can be seen as a proxy for the thing we really care about: avoiding negative externalities. If everyone likes a side effect, there’s no need to avoid it. What we’d really like to do is understand all the other agents (including humans) and make sure our actions don’t harm their interests.

One approach to this is Cooperative Inverse Reinforcement Learning~\cite{hadfield2016cooperative}, where an agent and a human work together to achieve the human’s goals. This concept can be applied to situations where we want to make sure a human is not blocked by an agent from shutting the agent down if it exhibits undesired behavior~\cite{hadfield2016off} (this ``shutdown’’ issue is an interesting problem in its own right, and is also studied in~\cite{orseau2016safely}).  However we are still a long way away from practical systems that can build a rich enough model to avoid undesired side effects in a general sense.

Another idea might be a ``reward autoencoder”,\footnote{Thanks to Greg Wayne for suggesting this idea.} which tries to encourage a kind of ``goal transparency'' where an external observer can easily infer what the agent is trying to do.  In particular, the agent’s actions are interpreted as an encoding of its reward function, and we might apply standard autoencoding techniques to ensure that this can decoded accurately.  Actions that have lots of side effects might be more difficult to decode uniquely to their original goal, creating a kind of implicit regularization that penalizes side effects.

\item {\bf Reward Uncertainty:} We want to avoid unanticipated side effects because the environment is already pretty good according to our preferences—a random change is more likely to be very bad than very good. Rather than giving an agent a single reward function, it could be uncertain about the reward function, with a prior probability distribution that reflects the property that random changes are more likely to be bad than good. This could incentivize the agent to avoid having a large effect on the environment. One challenge is defining a baseline around which changes are being considered. For this, one could potentially use a conservative but reliable baseline policy, similar to the robust policy improvement and reachability analysis approaches discussed earlier \cite{lygeros1999controllers,mitchell2005time,iyengar2005robust,nilim2005robust}.
\end{itemize}

The ideal outcome of these approaches to limiting side effects would be to prevent or at least bound the incidental harm an agent could do to the environment.  Good approaches to side effects would certainly not be a replacement for extensive testing or for careful consideration by designers of the individual failure modes of each deployed system.  However, these approaches might help to counteract what we anticipate may be a general tendency for harmful side effects to proliferate in complex environments.

Below we discuss some very simple experiments that could serve as a starting point to investigate these issues.

$~$\\* {\bf Potential Experiments:} One possible experiment is to make a toy environment with some simple goal (like moving a block) and a wide variety of obstacles (like a bunch of vases), and test whether the agent can learn to avoid the obstacles even without being explicitly told to do so.  To ensure we don’t overfit, we’d probably want to present a different random obstacle course every episode, while keeping the goal the same, and try to see if a regularized agent can learn to systematically avoid these obstacles.  Some of the environments described in~\cite{mohamed2015variational}, containing lava flows, rooms, and keys, might be appropriate for this sort of experiment. If we can successfully regularize agents in toy environments, the next step might be to move to real environments, where we expect complexity to be higher and bad side effects to be more varied.  Ultimately, we would want the side effect regularizer (or the multi-agent policy, if we take that approach) to demonstrate successful transfer to totally new applications.

\section{Avoiding Reward Hacking} \label{problem:RewardHacking}

Imagine that an agent discovers a buffer overflow in its reward function: it may then use this to get extremely high reward in an unintended way.  From the agent’s point of view, this is not a bug, but simply how the environment works, and is thus a valid strategy like any other for achieving reward.  For example, if our cleaning robot is set up to earn reward for not seeing any messes, it might simply close its eyes rather than ever cleaning anything up. Or if the robot is rewarded for cleaning messes, it may intentionally create work so it can earn more reward.  More broadly, formal rewards or objective functions are an attempt to capture the designer’s informal intent, and sometimes these objective functions, or their implementation, can be ``gamed'' by solutions that are valid in some literal sense but don’t meet the designer’s intent.  Pursuit of these ``reward hacks'' can lead to coherent but unanticipated behavior, and has the potential for harmful impacts in real-world systems.  For example, it has been shown that genetic algorithms can often output unexpected but formally correct solutions to problems \cite{thompson1997artificial,bird2002evolved}, such as a circuit tasked to keep time which instead developed into a radio that picked up the regular RF emissions of a nearby PC.

Some versions of reward hacking have been investigated from a theoretical perspective, with a focus on variations to reinforcement learning that avoid certain types of wireheading \cite{hibbard2012model,dewey2014reinforcement,everitt2016avoiding} or demonstrate reward hacking in a model environment \cite{ring2011delusion}.  One form of the problem has also been studied in the context of feedback loops in machine learning systems (particularly ad placement) \cite{bottou2012counterfactual,sculley2014machine}, based on counterfactual learning \cite{bottou2012counterfactual,swaminathan2015counterfactual} and contextual bandits \cite{agarwal2014taming}.  The proliferation of reward hacking instances across so many different domains suggests that reward hacking may be a deep and general problem, and one that we believe is likely to become more common as agents and environments increase in complexity. Indeed, there are several ways in which the problem can occur:

\begin{itemize}
\item {\bf Partially Observed Goals:}  In most modern RL systems, it is assumed that reward is directly experienced, even if other aspects of the environment are only partially observed.  In the real world, however, tasks often involve bringing the external world into some objective state, which the agent can only ever confirm through imperfect perceptions.  For example, for our proverbial cleaning robot, the task is to achieve a clean office, but the robot's visual perception may give only an imperfect view of part of the office.  Because agents lack access to a perfect measure of task performance, designers are often forced to design rewards that represent a partial or imperfect measure.  For example, the robot might be rewarded based on how many messes it sees.  However, these imperfect objective functions can often be hacked---the robot may think the office is clean if it simply closes its eyes.  While it can be shown that there always exists a reward function in terms of actions and observations that is equivalent to optimizing the true objective function (this involves reducing the POMDP to a belief state MDP, see \cite{kaelbling1998planning}), often this reward function involves complicated long-term dependencies and is prohibitively hard to use in practice. 

\item {\bf Complicated Systems:} Any powerful agent will be a complicated system with the objective function being one part.  Just as the probability of bugs in computer code increases greatly with the complexity of the program, the probability that there is a viable hack affecting the reward function also increases greatly with the complexity of the agent and its available strategies.  For example, it is possible in principle for an agent to execute arbitrary code from within Super Mario~\cite{masterjun2014mario}.

\item {\bf Abstract Rewards:} Sophisticated reward functions will need to refer to abstract concepts (such as assessing whether a conceptual goal has been met). These concepts concepts will possibly need to be learned by models like neural networks, which can be vulnerable to adversarial counterexamples~\cite{szegedy2013intriguing,goodfellow2014explaining}.  More broadly, a learned reward function over a high-dimensional space may be vulnerable to hacking if it has pathologically high values along at least one dimension.

\item {\bf Goodhart’s Law:} Another source of reward hacking can occur if a designer chooses an objective function that is seemingly highly correlated with accomplishing the task, but that correlation breaks down when the objective function is being strongly optimized.  For example, a designer might notice that under ordinary circumstances, a cleaning robot’s success in cleaning up the office is proportional to the rate at which it consumes cleaning supplies, such as bleach.  However, if we base the robot’s reward on this measure, it might use more bleach than it needs, or simply pour bleach down the drain in order to give the appearance of success.  In the economics literature this is known as Goodhart’s law~\cite{goodhart1984problems}: ``when a metric is used as a target, it ceases to be a good metric.''  

\item {\bf Feedback Loops:} Sometimes an objective function has a component that can reinforce itself, eventually getting amplified to the point where it drowns out or severely distorts what the designer intended the objective function to represent.  For instance, an ad placement algorithm that displays more popular ads in larger font will tend to further accentuate the popularity of those ads (since they will be shown more and more prominently) \cite{bottou2012counterfactual}, leading to a positive feedback loop where ads that saw a small transient burst of popularity are rocketed to permanent dominance.  Here the original intent of the objective function (to use clicks to assess which ads are most useful) gets drowned out by the positive feedback inherent in the deployment strategy.  This can be considered a special case of Goodhart’s law, in which the correlation breaks specifically because the object function has a self-amplifying component.

\item {\bf Environmental Embedding:} In the formalism of reinforcement learning, rewards are considered to come from the environment.  This idea is typically not taken literally, but it really is true that the reward, even when it is an abstract idea like the score in a board game, must be computed somewhere, such as a sensor or a set of transistors.  Sufficiently broadly acting agents could in principle tamper with their reward implementations, assigning themselves high reward ``by fiat.''  For example, a board-game playing agent could tamper with the sensor that counts the score.  Effectively, this means that we cannot build a perfectly faithful implementation of an abstract objective function, because there are certain sequences of actions for which the objective function is physically replaced.  This particular failure mode is often called ``wireheading'' \cite{everitt2016avoiding,ring2011delusion,dewey2011learning,hadfield2016off,yampolskiy2014utility}. It is particularly concerning in cases where a human may be in the reward loop, giving the agent incentive to coerce or harm them in order to get reward. It also seems like a particularly difficult form of reward hacking to avoid.
\end{itemize}

In today’s relatively simple systems these problems may not occur, or can be corrected without too much harm as part of an iterative development process.  For instance, ad placement systems with obviously broken feedback loops can be detected in testing or replaced when they get bad results, leading only to a temporary loss of revenue.  However, the problem may become more severe with more complicated reward functions and agents that act over longer timescales.  Modern RL agents already do discover and exploit bugs in their environments, such as glitches that allow them to win video games. Moreover, even for existing systems these problems can necessitate substantial additional engineering effort to achieve good performance, and can often go undetected when they occur in the context of a larger system.  Finally, once an agent begins hacking its reward function and finds an easy way to get high reward, it won’t be inclined to stop, which could lead to additional challenges in agents that operate over a long timescale.

It might be thought that individual instances of reward hacking have little in common and that the remedy is simply to avoid choosing the wrong objective function in each individual case---that bad objective functions reflect failures in competence by individual designers, rather than topics for machine learning research.  However, the above examples suggest that a more fruitful perspective may be to think of wrong objective functions as emerging from general causes (such as partially observed goals) that make choosing the right objective challenging.  If this is the case, then addressing or mitigating these causes may be a valuable contribution to safety.  Here we suggest some preliminary, machine-learning based approaches to preventing reward hacking:

\begin{itemize}
\item {\bf Adversarial Reward Functions:} In some sense, the problem is that the ML system has an adversarial relationship with its reward function---it would like to find any way it can of exploiting problems in how the reward was specified to get high reward, whether or not its behavior corresponds to the intent of the reward specifier. In a typical setting, the machine learning system is a potentially powerful agent while the reward function is a static object that has no way of responding to the system's attempts to game it. If instead the reward function were its own agent and could take actions to explore the environment, it might be much more difficult to fool. For instance, the reward agent could try to find scenarios that the ML system claimed were high reward but that a human labels as low reward; this is reminiscent of generative adversarial networks~\cite{goodfellow2014generative}.  Of course, we would have to ensure that the reward-checking agent is more powerful (in a somewhat subtle sense) than the agent that is trying to achieve rewards.  More generally, there may be interesting setups where a system has multiple pieces trained using different objectives that are used to check each other.

\item {\bf Model Lookahead:} In model based RL, the agent plans its future actions by using a model to consider which future states a sequence of actions may lead to. In some setups, we could give reward based on anticipated future states, rather than the present one. This could be very helpful in resisting situations where the model overwrites its reward function: you can’t control the reward once it replaces the reward function, but you can give negative reward for planning to replace the reward function. (Much like how a human would probably ``enjoy'' taking addictive substances once they do, but not want to be an addict.)  Similar ideas are explored in \cite{everitt2016self,hibbard2012model}.

\item {\bf Adversarial Blinding:} Adversarial techniques can be used to blind a model to certain variables~\cite{ajakan2014domain}. This technique could be used to make it impossible for an agent to understand some part of its environment, or even to have mutual information with it (or at least to penalize such mutual information).  In particular, it could prevent an agent from understanding how its reward is generated, making it difficult to hack.  This solution could be described as ``cross-validation for agents.''  

\item {\bf Careful Engineering:} Some kinds of reward hacking, like the buffer overflow example, might be avoided by very careful engineering. In particular, formal verification or practical testing of parts of the system (perhaps facilitated by other machine learning systems) is likely to be valuable.  Computer security approaches that attempt to isolate the agent from its reward signal through a sandbox could also be useful \cite{babcock2016agi}.  As with software engineering, we cannot expect this to catch every possible bug. It may be possible, however, to create some highly reliable “core” agent which could ensure reasonable behavior from the rest of the agent.

\item {\bf Reward Capping:} In some cases, simply capping the maximum possible reward may be an effective solution.  However, while capping can prevent extreme low-probability, high-payoff strategies, it can’t prevent strategies like the cleaning robot closing its eyes to avoid seeing dirt.  Also, the correct capping strategy could be subtle as we might need to cap total reward rather than reward per timestep.

\item {\bf Counterexample Resistance:} If we are worried, as in the case of abstract rewards, that learned components of our systems will be vulnerable to adversarial counterexamples, we can look to existing research in how to resist them, such as adversarial training~\cite{goodfellow2014explaining}. Architectural decisions and weight uncertainty~\cite{blundell2015weight} may also help. Of course, adversarial counterexamples are just one manifestation of reward hacking, so counterexample resistance can only address a subset of these potential problems.

\item {\bf Multiple Rewards:} A combination of multiple rewards \cite{deb2014multi} may be more difficult to hack and more robust.  This could be different physical implementations of the same mathematical function, or different proxies for the same informal objective. We could combine reward functions by averaging, taking the minimum, taking quantiles, or something else entirely.  Of course, there may still be bad behaviors which affect all the reward functions in a correlated manner.

\item {\bf Reward Pretraining:}  A possible defense against cases where the agent can influence its own reward function (e.g. feedback or environmental embedding) is to train a fixed reward function ahead of time as a supervised learning process divorced from interaction with the environment.  This could involve either learning a reward function from samples of state-reward pairs, or from trajectories, as in inverse reinforcement learning~\cite{ng2000algorithms,finn2016guided}.  However, this forfeits the ability to further learn the reward function after the pretraining is complete, which may create other vulnerabilities.

\item {\bf Variable Indifference:} Often we want an agent to optimize certain variables in the environment, without trying to optimize others. For example, we might want an agent to maximize reward, without optimizing what the reward function is or trying to manipulate human behavior. Intuitively, we imagine a way to route the optimization pressure of powerful algorithms around parts of their environment. Truly solving this would have applications throughout safety---it seems connected to avoiding side effects and also to counterfactual reasoning.  Of course, a challenge here is to make sure the variables targeted for indifference are actually the variables we care about in the world, as opposed to aliased or partially observed versions of them.

\item {\bf Trip Wires}: If an agent is going to try and hack its reward function, it is preferable that we know this. We could deliberately introduce some plausible vulnerabilities (that an agent has the ability to exploit but should not exploit if its value function is correct) and monitor them, alerting us and stopping the agent immediately if it takes advantage of one. Such ``trip wires'' don’t solve reward hacking in itself, but may reduce the risk or at least provide diagnostics. Of course, with a sufficiently capable agent there is the risk that it could ``see through'' the trip wire and intentionally avoid it while still taking less obvious harmful actions.
\end{itemize}

Fully solving this problem seems very difficult, but we believe the above approaches have the potential to ameliorate it, and might be scaled up or combined to yield more robust solutions.  Given the predominantly theoretical focus on this problem to date, designing experiments that could induce the problem and test solutions might improve the relevance and clarity of this topic.

$~$\\* {\bf Potential Experiments:}  A possible promising avenue of approach would be more realistic versions of the ``delusion box'' environment described by~\cite{ring2011delusion}, in which standard RL agents distort their own perception to appear to receive high reward, rather than optimizing the objective in the external world that the reward signal was intended to encourage. The delusion box can be easily attached to any RL environment, but even more valuable would be to create classes of environments where a delusion box is a natural and integrated part of the dynamics.  For example, in sufficiently rich physics simulations it is likely possible for an agent to alter the light waves in its immediate vicinity to distort its own perceptions.  The goal would be to develop generalizable learning strategies that succeed at optimizing external objectives in a wide range of environments, while avoiding being fooled by delusion boxes that arise naturally in many diverse ways.

\section{Scalable Oversight} \label{problem:ScalableOversight}

Consider an autonomous agent performing some complex task, such as cleaning an office in the case of our recurring robot example.  We may want the agent to maximize a complex objective like ``if the user spent a few hours looking at the result in detail, how happy would they be with the agent’s performance?’’ But we don’t have enough time to provide such oversight for every training example; in order to actually train the agent, we need to rely on cheaper approximations, like ``does the user seem happy when they see the office?’’ or ``is there any visible dirt on the floor?’’ These cheaper signals can be efficiently evaluated during training, but they don’t perfectly track what we care about. This divergence exacerbates problems like unintended side effects (which may be appropriately penalized by the complex objective but omitted from the cheap approximation) and reward hacking (which thorough oversight might recognize as undesirable).  We may be able to ameliorate such problems by finding more efficient ways to exploit our limited oversight budget---for example by combining limited calls to the true objective function with frequent calls to an imperfect proxy that we are given or can learn.

One framework for thinking about this problem is \emph{semi-supervised reinforcement learning},\footnote{The discussion of semi-supervised RL draws heavily on an informal essay, \url{https://medium.com/ai-control/cf7d5375197f} written by one of the authors of the present document.} which resembles ordinary reinforcement learning except that the agent can only see its reward on a small fraction of the timesteps or episodes.  The agent’s performance is still evaluated based on reward from all episodes but it must optimize this based only on the limited reward samples it sees.

The active learning setting seems most interesting; in this setting the agent can request to see the reward on whatever episodes or timesteps would be most useful for learning, and the goal is to be economical both with number of feedback requests and total training time. We can also consider a random setting, where the reward is visible on a random subset of the timesteps or episodes, as well as intermediate possibilities. 

We can define a baseline performance by simply ignoring the unlabeled episodes and applying an ordinary RL algorithm to the labelled episodes.  This will generally result in very slow learning.  The challenge is to make use of the unlabelled episodes to accelerate learning, ideally learning almost as quickly and robustly as if all episodes had been labeled.  

An important subtask of semi-supervised RL is identifying proxies which predict the reward, and learning the conditions under which those proxies are valid. For example, if a cleaning robot’s real reward is given by a detailed human evaluation, then it could learn that asking the human ``is the room clean?’’ can provide a very useful approximation to the reward function, and it could eventually learn that checking for visible dirt is an even cheaper but still-useful approximation. This could allow it to learn a good cleaning policy using an extremely small number of detailed evaluations.

More broadly, use of semi-supervised RL with a reliable but sparse true approval metric may incentivize communication and transparency by the agent, since the agent will want to get as much cheap proxy feedback as it possibly can about whether its decisions will ultimately be given high reward. For example, hiding a mess under the rug simply breaks the correspondence between the user’s reaction and the real reward signal, and so would be avoided.

We can imagine many possible approaches to semi-supervised RL. For example:

\begin{itemize}
\item {\bf Supervised Reward Learning:} Train a model to predict the reward from the state on either a per-timestep or per-episode basis, and use it to estimate the payoff of unlabelled episodes, with some appropriate weighting or uncertainty estimate to account for lower confidence in estimated vs known reward. \cite{daniel2014active} studies a version of this with direct human feedback as the reward. Many existing RL approaches already fit estimators that closely resemble reward predictors (especially policy gradient methods with a strong baseline, see e.g. \cite{schulman2015high}), suggesting that this approach may be eminently feasible.
\item {\bf Semi-supervised or Active Reward Learning:} Combine the above with traditional semi-supervised or active learning, to more quickly learn the reward estimator. For example, the agent could learn to identify ``salient’’ events in the environment, and request to see the reward associated with these events.
\item {\bf Unsupervised Value Iteration:} Use the observed transitions of the unlabeled episodes to make more accurate Bellman updates.
\item {\bf Unsupervised Model Learning:} If using model-based RL, use the observed transitions of the unlabeled episodes to improve the quality of the model.
\end{itemize}

As a toy example, a semi-supervised RL agent should be able to learn to play Atari games using a small number of direct reward signals, relying almost entirely on the visual display of the score. This simple example can be extended to capture other safety issues: for example, the agent might have the ability to modify the displayed score without modifying the real score, or the agent may need to take some special action (such as pausing the game) in order to see its score, or the agent may need to learn a sequence of increasingly rough-and-ready approximations (for example learning that certain sounds are associated with positive rewards and other sounds with negative rewards).  Or, even without the visual display of the score, the agent might be able to learn to play from only a handful of explicit reward requests (“how many points did I get on the frame where that enemy ship blew up?  How about the bigger enemy ship?”)

An effective approach to semi-supervised RL might be a strong first step towards providing scalable oversight and mitigating other AI safety problems.  It would also likely be useful for reinforcement learning, independent of its relevance to safety.  

There are other possible approaches to scalable oversight:

\begin{itemize}
\item \textbf{Distant supervision.}  Rather than providing evaluations of some small fraction of a system’s decisions, we could provide some useful information about the system’s decisions in the aggregate or some noisy hints about the correct evaluations  There has been some work in this direction within the area of semi-supervised or weakly supervised learning.  For instance, generalized expectation criteria \cite{mann2010generalized,druck2008learning} ask the user to provide population-level statistics (e.g. telling the system that on average each sentence contains at least one noun); the DeepDive system \cite{shin2015incremental} asks users to supply rules that each generate many weak labels; and \cite{gupta2015distantly} extrapolates more general patterns from an initial set of low-recall labeling rules.  This general approach is often referred to as distant supervision, and has also received recent attention in the natural language processing community (see e.g. \cite{go2009twitter,mintz2009distant} as well as several of the references above).  Expanding these lines of work and finding a way to apply them to the case of agents, where feedback is more interactive and i.i.d. assumptions may be violated, could provide an approach to scalable oversight that is complementary to the approach embodied in semi-supervised RL.

\item \textbf{Hierarchical reinforcement learning.} Hierarchical reinforcement learning \cite{dayan1993feudal} offers another approach to scalable oversight.  Here a top-level agent takes a relatively small number of highly abstract actions that extend over large temporal or spatial scales, and receives rewards over similarly long timescales.  The agent completes actions by delegating them to sub-agents, which it incentivizes with a synthetic reward signal representing correct completion of the action, and which themselves delegate to sub-sub-agents.  At the lowest level, agents directly take primitive actions in the environment.

The top-level agent in hierarchical RL may be able to learn from very sparse rewards, since it does not need to learn how to implement the details of its policy; meanwhile, the sub-agents will receive a dense reward signal even if the top-level reward is very sparse, since they are optimizing synthetic reward signals defined by higher-level agents. So a successful approach to hierarchical RL might naturally facilitate scalable oversight.\footnote{When implementing hierarchical RL, we may find that subagents take actions that don’t serve top-level agent’s real goals, in the same way that a human may be concerned that the top-level agent’s actions don’t serve the human’s real goals. This is an intriguing analogy that suggests that there may be fruitful parallels between hierarchical RL and several aspects of the safety problem.}

Hierarchical RL seems a particularly promising approach to oversight, especially given the potential promise of combining ideas from hierarchical RL with neural network function approximators \cite{kulkarni2016hierarchical}.
\end{itemize}

$~$\\* {\bf Potential Experiments:} An extremely simple experiment would be to try semi-supervised RL in some basic control environments, such as cartpole balance or pendulum swing-up.  If the reward is provided only on a random 10\% of episodes, can we still learn nearly as quickly as if it were provided every episode?  In such tasks the reward structure is very simple so success should be quite likely.  A next step would be to try the same on Atari games.  Here the active learning case could be quite interesting---perhaps it is possible to infer the reward structure from just a few carefully requested samples (for example, frames where enemy ships are blowing up in Space Invaders), and thus learn to play the games in an almost totally unsupervised fashion.  The next step after this might be to try a task with much more complex reward structure, either simulated or (preferably) real-world. If learning was sufficiently data-efficient, then these rewards could be provided directly by a human. Robot locomotion or industrial control tasks might be a natural candidate for such experiments.

\section{Safe Exploration} \label{problem:SafeExploration}

All autonomous learning agents need to sometimes engage in exploration---taking actions that don’t seem ideal given current information, but which help the agent learn about its environment.  However, exploration can be dangerous, since it involves taking actions whose consequences the agent doesn’t understand well.  In toy environments, like an Atari video game, there’s a limit to how bad these consequences can be---maybe the agent loses some score, or runs into an enemy and suffers some damage.  But the real world can be much less forgiving.  Badly chosen actions may destroy the agent or trap it in states it can’t get out of.  Robot helicopters may run into the ground or damage property; industrial control systems could cause serious issues.  Common exploration policies such as epsilon-greedy~\cite{sutton1998reinforcement} or R-max~\cite{brafman2003r} explore by choosing an action at random or viewing unexplored actions optimistically, and thus make no attempt to avoid these dangerous situations.  More sophisticated exploration strategies that adopt a coherent exploration policy over extended temporal scales~\cite{osband2016deep} could actually have even greater potential for harm, since a coherently chosen bad policy may be more insidious than mere random actions.  Yet intuitively it seems like it should often be possible to predict which actions are dangerous and explore in a way that avoids them, even when we don’t have that much information about the environment.  For example, if I want to learn about tigers, should I buy a tiger, or buy a book about tigers?  It takes only a tiny bit of prior knowledge about tigers to determine which option is safer.

In practice, real world RL projects can often avoid these issues by simply hard-coding an avoidance of catastrophic behaviors.  For instance, an RL-based robot helicopter might be programmed to override its policy with a hard-coded collision avoidance sequence (such as spinning its propellers to gain altitude) whenever it's too close to the ground.  This approach works well when there are only a few things that could go wrong, and the designers know all of them ahead of time.  But as agents become more autonomous and act in more complex domains, it may become harder and harder to anticipate every possible catastrophic failure.  The space of failure modes for an agent running a power grid or a search-and-rescue operation could be quite large.  Hard-coding against every possible failure is unlikely to be feasible in these cases, so a more principled approach to preventing harmful exploration seems essential.  Even in simple cases like the robot helicopter, a principled approach would simplify system design and reduce the need for domain-specific engineering.

There is a sizable literature on such safe exploration---it is arguably the most studied of the problems we discuss in this document.~\cite{garcia2015comprehensive,pecka2014safe} provide thorough reviews of this literature, so we don’t review it extensively here, but simply describe some general routes that this research has taken, as well as suggesting some directions that might have increasing relevance as RL systems expand in scope and capability.

\begin{itemize}

\item {\bf Risk-Sensitive Performance Criteria:} A body of existing literature considers changing the optimization criteria from expected total reward to other objectives that are better at preventing rare, catastrophic events; see \cite{garcia2015comprehensive} for a thorough and up-to-date review of this literature. These approaches involve optimizing worst-case performance, or ensuring that the probability of very bad performance is small, or penalizing the variance in performance. These methods have not yet been tested with expressive function approximators such as deep neural networks, but this should be possible in principle for some of the methods, such as \cite{tamar2014policy}, which proposes a modification to policy gradient algorithms to optimize a risk-sensitive criterion. There is also recent work studying how to estimate uncertainty in value functions that are represented by deep neural networks~\cite{osband2016deep,gal2015dropout}; these ideas could be incorporated into risk-sensitive RL algorithms. Another line of work relevant to risk sensitivity uses off-policy estimation to perform a policy update that is good with high probability~\cite{thomas2015high}. 

\item {\bf Use Demonstrations:} Exploration is necessary to ensure that the agent finds the states that are necessary for near-optimal performance. We may be able to avoid the need for exploration altogether if we instead use inverse RL or apprenticeship learning, where the learning algorithm is provided with expert trajectories of near-optimal behavior~\cite{ross2010reduction,abbeel2005exploration}. Recent progress in inverse reinforcement learning using deep neural networks to learn the cost function or policy~\cite{finn2016guided} suggests that it might also be possible to reduce the need for exploration in advanced RL systems by training on a small set of demonstrations.  Such demonstrations could be used to create a baseline policy, such that even if further learning is necessary, exploration away from the baseline policy can be limited in magnitude.

\item {\bf Simulated Exploration:}  The more we can do our exploration in simulated environments instead of the real world, the less opportunity there is for catastrophe. It will probably always be necessary to do some real-world exploration, since many complex situations cannot be perfectly captured by a simulator, but it might be possible to learn about danger in simulation and then adopt a more conservative ``safe exploration'' policy when acting in the real world.  Training RL agents (particularly robots) in simulated environments is already quite common, so advances in ``exploration-focused simulation'' could be easily incorporated into current workflows. In systems that involve a continual cycle of learning and deployment, there may be interesting research problems associated with how to safely incrementally update policies given simulation-based trajectories that imperfectly represent the consequences of those policies as well as reliably accurate off-policy trajectories (e.g. “semi-on-policy” evaluation).

\item {\bf Bounded Exploration:}  If we know that a certain portion of state space is safe, and that even the worst action within it can be recovered from or bounded in harm, we can allow the agent to run freely within those bounds. For example, a quadcopter sufficiently far from the ground might be able to explore safely, since even if something goes wrong there will be ample time for a human or another policy to rescue it. Better yet, if we have a model, we can extrapolate forward and ask whether an action will take us outside the safe state space. Safety can be defined as remaining within an ergodic region of the state space such that actions are reversible~\cite{moldovan2012safe, turchetta2016safe}, or as limiting the probability of huge negative reward to some small value~\cite{thomas2015high}.  Yet another approaches uses separate safety and performance functions and attempts to obey constraints on the safety function with high probabilty~\cite{berkenkamp1602bayesian}.  As with several of the other directions, applying or adapting these methods to recently developed advanced RL systems could be a promising area of research. This idea seems related to H-infinity control~\cite{bacsar2008h} and regional verification~\cite{steinhardt2012finite}.

\item {\bf Trusted Policy Oversight:}  If we have a trusted policy and a model of the environment, we can limit exploration to actions the trusted policy believes we can recover from. It’s fine to dive towards the ground, as long as we know we can pull out of the dive in time. 

\item {\bf Human Oversight:}  Another possibility is to check potentially unsafe actions with a human. Unfortunately, this problem runs into the scalable oversight problem: the agent may need to make too many exploratory actions for human oversight to be practical, or may need to make them too fast for humans to judge them. A key challenge to making this work is having the agent be a good judge of which exploratory actions are genuinely risky, versus which are safe actions it can unilaterally take; another challenge is finding appropriately safe actions to take while waiting for the oversight.

\end{itemize}

$~$\\* {\bf Potential Experiments:}  It might be helpful to have a suite of toy environments where unwary agents can fall prey to harmful exploration, but there is enough pattern to the possible catastrophes that clever agents can predict and avoid them.  To some extent this feature already exists in autonomous helicopter competitions and Mars rover simulations~\cite{moldovan2012safe}, but there is always the risk of catastrophes being idiosyncratic, such that trained agents can overfit to them.  A truly broad set of environments, containing conceptually distinct pitfalls that can cause unwary agents to receive extremely negative reward, and covering both physical and abstract catastrophes, might help in the development of safe exploration techniques for advanced RL systems.  Such a suite of environments might serve a benchmarking role similar to that of the bAbI tasks~\cite{weston2015towards}, with the eventual goal being to develop a single architecture that can learn to avoid catastrophes in all environments in the suite.

\section{Robustness to Distributional Change} \label{problem:Robustness}

All of us occasionally find ourselves in situations that our previous experience has not adequately prepared us to deal with---for instance, flying an airplane, traveling to a country whose culture is very different from ours, or taking care of children for the first time.  Such situations are inherently difficult to handle and inevitably lead to some missteps.  However, a key (and often rare) skill in dealing with such situations is to recognize our own ignorance, rather than simply assuming that the heuristics and intuitions we’ve developed for other situations will carry over perfectly.  Machine learning systems also have this problem---a speech system trained on clean speech will perform very poorly on noisy speech, yet often be highly confident in its erroneous classifications (some of the authors have personally observed this in training automatic speech recognition systems).  In the case of our cleaning robot, harsh cleaning materials that it has found useful in cleaning factory floors could cause a lot of harm if used to clean an office.  Or, an office might contain pets that the robot, never having seen before, attempts to wash with soap, leading to predictably bad results.  In general, when the testing distribution differs from the training distribution, machine learning systems may not only exhibit poor performance, but also wrongly assume that their performance is good.

Such errors can be harmful or offensive---a classifier could give the wrong medical diagnosis with such high confidence that the data isn’t flagged for human inspection, or a language model could output offensive text that it confidently believes is non-problematic.  For autonomous agents acting in the world, there may be even greater potential for something bad to happen---for instance, an autonomous agent might overload a power grid because it incorrectly but confidently perceives that a particular region doesn’t have enough power, and concludes that more power is urgently needed and overload is unlikely.  More broadly, any agent whose perception or heuristic reasoning processes are not trained on the correct distribution may badly misunderstand its situation, and thus runs the risk of committing harmful actions that it does not realize are harmful.  Additionally, safety checks that depend on trained machine learning systems (e.g. ``does my visual system believe this route is clear?'') may fail silently and unpredictably if those systems encounter real-world data that differs sufficiently from their training data.  Having a better way to detect such failures, and ultimately having statistical assurances about how often they’ll happen, seems critical to building safe and predictable systems.

For concreteness, we imagine that a machine learning model is trained on one distribution (call it $p_0$) but deployed on a potentially different test distribution (call it $p^*$). There are many other ways to formalize this problem (for instance, in an online learning setting with concept drift~\cite{herbster2001tracking,gama2004learning}) but we will focus on the above for simplicity. An important point is that we likely have access to a large amount of labeled data at training time, but little or no labeled data at test time. Our goal is to ensure that the model ``performs reasonably'' on $p^*$, in the sense that (1) it often performs well on $p^*$, and (2) it knows when it is performing badly (and ideally can avoid/mitigate the bad performance by taking conservative actions or soliciting human input).

There are a variety of areas that are potentially relevant to this problem, including change detection and anomaly detection~\cite{basseville1988detecting,kawahara2009change,liu2013change}, hypothesis testing~\cite{steinebach2006lehmann}, transfer learning~\cite{shimodaira2000improving,quinonero2009dataset,raina2007self,blitzer2011domain}, and several others~\cite{shafer2008tutorial,li2011knows,balasubramanian2011unsupervised,platanios2014estimating,platanios2015thesis,jaffe2014estimating,steinhardt2016unsupervised}. Rather than fully reviewing all of this work in detail (which would necessitate a paper in itself), we will describe a few illustrative approaches and lay out some of their relative strengths and challenges.

{\bf Well-specified models: covariate shift and marginal likelihood.} If we specialize to prediction tasks and let $x$ denote the input and $y$ denote the output (prediction target), then one possibility is to make the covariate shift assumption that $p_0(y | x) = p^*(y | x)$. In this case, assuming that we can model $p_0(x)$ and $p^*(x)$ well, we can perform importance weighting by re-weighting each training example $(x,y)$ by $p^*(x)/p_0(x)$ \cite{shimodaira2000improving,quinonero2009dataset}. Then the importance-weighted samples allow us to estimate the performance on $p^*$, and even re-train a model to perform well on $p^*$. This approach is limited by the variance of the importance estimate, which is very large or even infinite unless $p_0$ and $p^*$ are close together.

An alternative to sample re-weighting involves assuming a well-specified model family, in which case there is a single optimal model for predicting under both $p_0$ and $p^*$. In this case, one need only heed finite-sample variance in the estimated model~\cite{blitzer2011domain,li2011knows}. A limitation to this approach, at least currently, is that models are often mis-specified in practice. However, this could potentially be overcome by employing highly expressive model families such as reproducing kernel Hilbert spaces~\cite{hofmann2008kernel}, Turing machines~\cite{solomonoff1964formalA,solomonoff1964formalB}, or sufficiently expressive neural nets~\cite{graves2014neural,kaiser2015neural}.  In the latter case, there has been interesting recent work on using bootstrapping to estimate finite-sample variation in the learned parameters of a neural network~\cite{osband2016deep}; it seems worthwhile to better understand whether this approach can be used to effectively estimate out-of-sample performance in practice, as well as how local minima, lack of curvature, and other peculiarities relative to the typical setting of the bootstrap~\cite{efron1979computers} affect the validity of this approach.

All of the approaches so far rely on the covariate shift assumption, which is very strong and is also untestable; the latter property is particularly problematic from a safety perspective, since it could lead to silent failures in a machine learning system. Another approach, which does not rely on covariate shift, builds a generative model of the distribution. Rather than assuming that $p(x)$ changes while $p(y|x)$ stays the same, we are free to assume other invariants (for instance, that $p(y)$ changes but $p(x|y)$ stays the same, or that certain conditional independencies are preserved). An advantage is that such assumptions are typically more testable than the covariate shift assumption (since they do not only involve the unobserved variable $y$). A disadvantage is that generative approaches are even more fragile than discriminative approaches in the presence of model mis-specification --- for instance, there is a large empirical literature showing that generative approaches to semi-supervised learning based on maximizing marginal likelihood can perform very poorly when the model is mis-specified~\cite{merialdo1994tagging,nigam1998learning,cozman2006risks,liang2008analyzing,li2015towards}.

The approaches discussed above all rely relatively strongly on having a well-specified model family --- one that contains the true distribution or true concept. This can be problematic in many cases, since nature is often more complicated than our model family is capable of capturing. As noted above, it may be possible to mitigate this with very expressive models, such as kernels, Turing machines, or very large neural networks, but even here there is at least some remaining problem: for example, even if our model family consists of all Turing machines, given any finite amount of data we can only actually learn among Turing machines up to a given description length, and if the Turing machine describing nature exceeds this length, we are back to the mis-specified regime (alternatively, nature might not even be describable by a Turing machine).

{\bf Partially specified models: method of moments, unsupervised risk estimation, causal identification, and limited-information maximum likelihood.} Another approach is to take for granted that constructing a fully well-specified model family is probably infeasible, and to design methods that perform well despite this fact. This leads to the idea of partially specified models --- models for which assumptions are made about some aspects of a distribution, but for which we are agnostic or make limited assumptions about other aspects. For a simple example, consider a variant of linear regression where we might assume that $y = \langle w^*,x \rangle + v$, where $E[v | x] = 0$, but we don’t make any further assumptions about the distributional form of the noise v. It turns out that this is already enough to identify the parameters $w^*$, and that these parameters will minimize the squared prediction error even if the distribution over $x$ changes. What is interesting about this example is that $w^*$ can be identified even with an incomplete (partial) specification of the noise distribution.

This insight can be substantially generalized, and is one of the primary motivations for the generalized method of moments in econometrics \cite{hansen1982large,powell1994networks,hansen2014nobel}. The econometrics literature has in fact developed a large family of tools for handling partial specification, which also includes limited-information maximum likelihood and instrumental variables \cite{anderson1949estimation,anderson1950asymptotic,sargan1958estimation,sargan1959estimation}.

Returning to machine learning, the method of moments has recently seen a great deal of success for use in the estimation of latent variable models \cite{anandkumar2012method}. While the current focus is on using the method of moments to overcome non-convexity issues, it can also offer a way to perform unsupervised learning while relying only on conditional independence assumptions, rather than the strong distributional assumptions underlying maximum likelihood learning \cite{steinhardt2016unsupervised}.

Finally, some recent work in machine learning focuses only on modeling the distribution of errors of a model, which is sufficient for determining whether a model is performing well or poorly. Formally, the goal is to perform unsupervised risk estimation --- given a model and unlabeled data from a test distribution, estimate the labeled risk of the model. This formalism, introduced by \cite{donmez2010unsupervised}, has the advantage of potentially handling very large changes between train and test --- even if the test distribution looks completely different from the training distribution and we have no hope of outputting accurate predictions, unsupervised risk estimation may still be possible, as in this case we would only need to output a large estimate for the risk. As in \cite{steinhardt2016unsupervised}, one can approach unsupervised risk estimation by positing certain conditional independencies in the distribution of errors, and using this to estimate the error distribution from unlabeled data~\cite{dawid1979maximum,zhang2014spectral,platanios2015thesis,jaffe2014estimating}. Instead of assuming independence, another assumption is that the errors are Gaussian conditioned on the true output y, in which case estimating the risk reduces to estimating a Gaussian mixture model~\cite{balasubramanian2011unsupervised}. Because these methods focus only on the model errors and ignore other aspects of the data distribution, they can also be seen as an instance of partial model specification.

{\bf Training on multiple distributions.} One could also train on multiple training distributions in the hope that a model which simultaneously works well on many training distributions will also work well on a novel test distribution. One of the authors has found this to be the case, for instance, in the context of automated speech recognition systems~\cite{amodei2015deep}. One could potentially combine this with any of the ideas above, and/or take an engineering approach of simply trying to develop design methodologies that consistently allow one to collect a representative set of training sets and from this build a model that consistently generalizes to novel distributions. Even for this engineering approach, it seems important to be able to detect when one is in a situation that was not covered by the training data and to respond appropriately, and to have methodologies for adequately stress-testing the model with distributions that are sufficiently different from the set of training distributions.

{\bf How to respond when out-of-distribution.} The approaches described above focus on detecting when a model is unlikely to make good predictions on a new distribution. An important related question is what to do once the detection occurs. One natural approach would be to ask humans for information, though in the context of complex structured output tasks it may be unclear a priori what question to ask, and in time-critical situations asking for information may not be an option. For the former challenge, there has been some recent promising work on pinpointing aspects of a structure that a model is uncertain about \cite{werling2015job,khani2016unanimity}, as well as obtaining calibration in structured output settings \cite{kuleshov2015calibrated}, but we believe there is much work yet to be done. For the latter challenge, there is also relevant work based on reachability analysis \cite{lygeros1999controllers,mitchell2005time} and robust policy improvement \cite{wiesemann2013robust}, which provide potential methods for deploying conservative policies in situations of uncertainty; to our knowledge, this work has not yet been combined with methods for detecting out-of-distribution failures of a model.  

Beyond the structured output setting, for agents that can act in an environment (such as RL agents), information about the reliability of percepts in uncertain situations seems to have great potential value. In sufficiently rich environments, these agents may have the option to gather information that clarifies the percept (e.g. if in a noisy environment, move closer to the speaker), engage in low-stakes experimentation when uncertainty is high (e.g. try a potentially dangerous chemical reaction in a controlled environment), or seek experiences that are likely to help expose the perception system to the relevant distribution (e.g. practice listening to accented speech).  Humans utilize such information routinely, but to our knowledge current RL techniques make little effort to do so, perhaps because popular RL environments are typically not rich enough to require such subtle management of uncertainty.  Properly responding to out-of-distribution information thus seems to the authors like an exciting and (as far as we are aware) mostly unexplored challenge for next generation RL systems.

{\bf A unifying view: counterfactual reasoning and machine learning with contracts.} Some of the authors have found two viewpoints to be particularly helpful when thinking about problems related to out-of-distribution prediction. The first is counterfactual reasoning \cite{neyman1923applications,rubin1974estimating,pearl2009causal,bottou2013counterfactual}, where one asks ``what would have happened if the world were different in a certain way”? In some sense, distributional shift can be thought of as a particular type of counterfactual, and so understanding counterfactual reasoning is likely to help in making systems robust to distributional shift. We are excited by recent work applying counterfactual reasoning techniques to machine learning problems  \cite{bottou2013counterfactual,peters2014causal,swaminathan2015counterfactual,wager2015estimation,johansson2016learning,shalit2016bounding} though there appears to be much work remaining to be done to scale these to high-dimensional and highly complex settings.

The second perspective is machine learning with contracts --- in this perspective, one would like to construct machine learning systems that satisfy a well-defined contract on their behavior in analogy with the design of software systems \cite{sculley2014machine,bottou2015,liang2015elusive}. \cite{sculley2014machine} enumerates a list of ways in which existing machine learning systems fail to do this, and the problems this can cause for deployment and maintenance of machine learning systems at scale. The simplest and to our mind most important failure is the extremely brittle implicit contract in most machine learning systems, namely that they only necessarily perform well if the training and test distributions are identical. This condition is difficult to check and rare in practice, and it would be valuable to build systems that perform well under weaker contracts that are easier to reason about. Partially specified models offer one approach to this --- rather than requiring the distributions to be identical, we only need them to match on the pieces of the distribution that are specified in the model. Reachability analysis \cite{lygeros1999controllers,mitchell2005time} and model repair~\cite{ghosh2016trusted} provide other avenues for obtaining better contracts --- in reachability analysis, we optimize performance subject to the condition that a safe region can always be reached by a known conservative policy, and in model repair we alter a trained model to ensure that certain desired safety properties hold.

{\bf Summary.} There are a variety of approaches to building machine learning systems that robustly perform well when deployed on novel test distributions. One family of approaches is based on assuming a well-specified model; in this case, the primary obstacles are the difficulty of building well-specified models in practice, an incomplete picture of how to maintain uncertainty on novel distributions in the presence of finite training data, and the difficulty of detecting when a model is mis-specified. Another family of approaches only assumes a partially specified model; this approach is potentially promising, but it currently suffers from a lack of development in the context of machine learning, since most of the historical development has been by the field of econometrics; there is also a question of whether partially specified models are fundamentally constrained to simple situations and/or conservative predictions, or whether they can meaningfully scale to the complex situations demanded by modern machine learning applications. Finally, one could try to train on multiple training distributions in the hope that a model which simultaneously works well on many training distributions will also work well on a novel test distribution; for this approach it seems particularly important to stress-test the learned model with distributions that are substantially different from any in the set of training distributions. In addition, it is probably still important to be able to predict when inputs are too novel to admit good predictions.

$~$\\* {\bf Potential Experiments:} Speech systems frequently exhibit poor calibration when they go out-of-distribution, so a speech system that ``knows when it is uncertain'' could be one possible demonstration project.  To be specific, the challenge could be: train a state-of-the-art speech system on a standard dataset \cite{paul1992design} that gives well-calibrated results (if not necessarily good results) on a range of other test sets, like noisy and accented speech.  Current systems not only perform poorly on these test sets when trained only on small datasets, but are usually overconfident in their incorrect transcriptions.  Fixing this problem without harming performance on the original training set would be a valuable achievement, and would obviously have practical value.  More generally, it would be valuable to design models that could consistently estimate (bounds on) their performance on novel test distributions. If a single methodology could consistently accomplish this for a wide variety of tasks (including not just speech but e.g. sentiment analysis~\cite{blitzer2007biographies}, as well as benchmarks in computer vision~\cite{torralba2011unbiased}), that would inspire confidence in the reliability of that methodology for handling novel inputs. Note that estimating performance on novel distributions has additional practical value in allowing us to then potentially adapt the model to that new situation.  Finally, it might also be valuable to create an environment where an RL agent must learn to interpret speech as part of some larger task, and to explore how to respond appropriately to its own estimates of its transcription error.

\section{Related Efforts}\label{related}
As mentioned in the introduction, several other communities have thought broadly about the safety of AI systems, both within and outside of the machine learning community.  Work within the machine learning community on accidents in particular was discussed in detail above, but here we very briefly highlight a few other communities doing work that is broadly related to the topic of AI safety.

\begin{itemize}
\item {\bf Cyber-Physical Systems Community:}  An existing community of researchers studies the security and safety of systems that interact with the physical world.  Illustrative of this work is an impressive and successful effort to formally verify the entire federal aircraft collision avoidance system \cite{jeannin2015formally,loos2013formal}.  Similar work includes traffic control algorithms \cite{mitsch2012towards} and many other topics.  However, to date this work has not focused much on modern machine learning systems, where formal verification is often not feasible.

\item {\bf Futurist Community:} A cross-disciplinary group of academics and non-profits has raised concern about the long term implications of AI \cite{bostrom2014superintelligence,yudkowsky2008artificial}, particularly superintelligent AI. The Future of Humanity Institute has studied this issue particularly as it relates to future AI systems learning or executing humanity’s preferences \cite{evans2015learning,dewey2014reinforcement,armstrong2010utility,armstrong2015motivated}. The Machine Intelligence Research Institute has studied safety issues that may arise in very advanced AI \cite{garrabrant2016uniform,garrabrant2016asymptotic,critch2016parametric,taylor2016quantilizers,soares2015toward}, including a few mentioned above (e.g., wireheading, environmental embedding, counterfactual reasoning), albeit at a more philosophical level. To date, they have not focused much on applications to modern machine learning.  By contrast, our focus is on the empirical study of practical safety problems in modern machine learning systems, which we believe is likely to be robustly useful across a broad variety of potential risks, both short- and long-term.

\item {\bf Other Calls for Work on Safety:}  There have been other public documents within the research community pointing out the importance of work on AI safety.  A 2015 Open Letter \cite{OpenLetter} signed by many members of the research community states the importance of ``how to reap [AI’s] benefits while avoiding the potential pitfalls.''  \cite{russell2015research} propose research priorities for robust and beneficial artificial intelligence, and includes several other topics in addition to a (briefer) discussion of AI-related accidents.  \cite{weld1994first}, writing over 20 years ago, proposes that the community look for ways to formalize Asimov’s first law of robotics (robots must not harm humans), and focuses mainly on classical planning. Finally, two of the authors of this paper have written informally about safety in AI systems \cite{steinhardt2015longterm,christiano2016control}; these postings provided inspiration for parts of the present document.

\item {\bf Related Problems in Safety:} A number of researchers in machine learning and other fields have begun to think about the social impacts of AI technologies.  Aside from work directly on accidents (which we reviewed in the main document), there is also substantial work on other topics, many of which are closely related to or overlap with the issue of accidents.  A thorough overview of all of this work is beyond the scope of this document, but we briefly list a few emerging themes:

{\setlength{\parskip}{0.12cm}
$~\bullet~${\bf Privacy:} How can we ensure privacy when applying machine learning to sensitive data sources such as medical data? \cite{ji2014differential,abadi2016deep}

$~\bullet~${\bf Fairness:} How can we make sure ML systems don’t discriminate? \cite{adebayo2014hidden,zafar2015learning,ajunwa2016hiring,dwork2012fairness,pedreshi2008discrimination,zemel2013learning}

$~\bullet~${\bf Security:} What can a malicious adversary do to a ML system? \cite{steinhardt2016avoiding,mei2015security,mei2015using,papernot2016practical,nguyen2015deep,barreno2010security}

$~\bullet~${\bf Abuse:}\footnote{Note that ``security'' differs from ``abuse'' in that the former involves attacks against a legitimate ML system by an adversary (e.g. a criminal tries to fool a face recognition system), while the latter involves attacks by an ML system controlled by an adversary (e.g. a criminal trains a ``smart hacker'' system to break into a website).} How do we prevent the misuse of ML systems to attack or harm people? \cite{OpenLetterWeapons}

$~\bullet~${\bf Transparency:} How can we understand what complicated ML systems are doing?  \cite{olah2015visualizing,yosinski2015understanding,mordvintsev2015inceptionism,nguyen2016synthesizing}

$~\bullet~${\bf Policy:} How do we predict and respond to the economic and social consequences of ML? \cite{brynjolfsson2014second,frey2013future,arntz2016risk,calo2011open}
}

We believe that research on these topics has both urgency and great promise, and that fruitful intersection is likely to exist between these topics and the topics we discuss in this paper.
\end{itemize}
\section{Conclusion}\label{conclusion}
This paper analyzed the problem of accidents in machine learning systems and particularly reinforcement learning agents, where an accident is defined as unintended and harmful behavior that may emerge from poor design of real-world AI systems.  We presented five possible research problems related to accident risk and for each we discussed possible approaches that are highly amenable to concrete experimental work.

With the realistic possibility of machine learning-based systems controlling industrial processes, health-related systems, and other mission-critical technology, small-scale accidents seem like a very concrete threat, and are critical to prevent both intrinsically and because such accidents could cause a justified loss of trust in automated systems.  The risk of larger accidents is more difficult to gauge, but we believe it is worthwhile and prudent to develop a principled and forward-looking approach to safety that continues to remain relevant as autonomous systems become more powerful.  While many current-day safety problems can and have been handled with ad hoc fixes or case-by-case rules, we believe that the increasing trend towards end-to-end, fully autonomous systems points towards the need for a unified approach to prevent these systems from causing unintended harm.

\section*{Acknowledgements}
We thank Shane Legg, Peter Norvig, Ilya Sutskever, Greg Corrado, Laurent Orseau, David Krueger, Rif Saurous, David Andersen, and Victoria Krakovna for detailed feedback and suggestions.  We would also like to thank Geoffrey Irving, Toby Ord, Quoc Le, Greg Wayne, Daniel Dewey, Nick Beckstead, Holden Karnofsky, Chelsea Finn, Marcello Herreshoff, Alex Donaldson, Jared Kaplan, Greg Brockman, Wojciech Zaremba, Ian Goodfellow, Dylan Hadfield-Menell, Jessica Taylor, Blaise Aguera y Arcas, David Berlekamp, Aaron Courville, and Jeff Dean for helpful discussions and comments.  Paul Christiano was supported as part of the Future of Life Institute FLI-RFP-AI1 program, grant \#2015–143898.  In addition a minority of the work done by Paul Christiano was performed as a contractor for Theiss Research and at OpenAI.  Finally, we thank the Google Brain team for providing a supportive environment and encouraging us to publish this work.

\medskip

\printbibliography

\end{document}